\def\year{2022}\relax
\documentclass[letterpaper]{article} 
\usepackage{aaai22}  
\usepackage{times}  
\usepackage{helvet}  
\usepackage{courier}  
\usepackage[hyphens]{url}  
\usepackage{graphicx} 
\usepackage{natbib}  
\usepackage{caption} 
\DeclareCaptionStyle{ruled}{labelfont=normalfont,labelsep=colon,strut=off} 
\usepackage{algorithm}
\usepackage{algorithmic}

\usepackage{microtype}
\usepackage{amssymb}
\usepackage{amsmath}
\usepackage{dsfont}
\usepackage{subfigure}

\usepackage{newfloat}
\usepackage{listings}
\floatstyle{ruled}
\newfloat{listing}{tb}{lst}{}
\floatname{listing}{Listing}
\title{A Semi-supervised Learning Approach with Two Teachers \\ to Improve Breakdown Identification in Dialogues}
\author{Qian Lin, Hwee Tou Ng
}
\affiliations{
Department of Computer Science, National University of Singapore\\
qlin@u.nus.edu, nght@comp.nus.edu.sg
}

\usepackage{bibentry}

\begin{document}

\maketitle

\begin{abstract}

Identifying breakdowns in ongoing dialogues helps to improve communication effectiveness. Most prior work on this topic relies on human annotated data and data augmentation to learn a classification model. While quality labeled dialogue data requires human annotation and is usually expensive to obtain, unlabeled data is easier to collect from various sources. In this paper, we propose a novel semi-supervised teacher-student learning framework to tackle this task. We introduce two teachers which are trained on labeled data and perturbed labeled data respectively. We leverage unlabeled data to improve classification in student training where we employ two teachers to refine the labeling of unlabeled data through teacher-student learning in a bootstrapping manner. Through our proposed training approach, the student can achieve improvements over single-teacher performance. Experimental results on the Dialogue Breakdown Detection Challenge dataset DBDC5 and Learning to Identify Follow-Up Questions dataset LIF show that our approach outperforms all previous published approaches as well as other supervised and semi-supervised baseline methods.

\end{abstract}

\section{Introduction}
In recent years, interactive virtual conversational agents have been developed rapidly and used widely in daily lives.
The information exchange between a user and an agent is done via a conversational dialogue.
To achieve effective communication, the agent is expected to generate a proper and rational response based on not only the last turn but also all previous utterances in the dialogue history to continue the dialogue.
The user's trust in the agent is damaged when the agent fails to identify the user's intent and generates an inappropriate response, which confuses the user and causes a breakdown in the dialogue.
Therefore, identifying breakdowns in dialogues is essential for improving the effectiveness of conversational agents, so that the agent is able to avoid generating responses which cause the breakdowns.

Much prior work on breakdown identification in dialogues has focused on supervised learning on human annotated data.
One line of work relies on feature-engineered machine learning methods including decision trees and random forests~\citep{wang-2019-dbdc4-waseda-tree-lstm}.
Another line of work utilizes non-Transformer based neural networks such as LSTM~\citep{hendriksen-2019-dbdc4-lstm,wang-2019-dbdc4-waseda-tree-lstm,shin-2019-dbdc4-memnet-bilstm}.
Transformer-based methods involve pre-trained language models which are pre-trained on large corpora~\citep{devlin-etal-2019-bert,conneau-etal-2020-xlmr}.
\citet{sugiyama-2019-dbdc4-bert} and utilize BERT with input consisting of the text and textual features from the dialogue. \citet{lin-etal-2020-co} introduce multilingual transfer learning through a cross-lingual pre-trained language model and co-attention modules to reason between the dialogue history and the last utterance.

A recent work~\citep{ng-2020-improving} proposes to perform pre-training on BERT with conversational data and apply self-supervised data augmentation on labeled data.
Although good performance has been reported, we observe that the gain of either continued pre-training or data augmentation on labeled data is marginal over the conventional BERT classification scheme.
Moreover, pre-training of a pre-trained language model on large corpora is resource-intensive.

We believe that training with dialogue data from other sources introduces diversity and enables the trained model to generalize better.
Since annotated dialogue data is expensive to obtain, we propose using unlabeled data through semi-supervised learning and self-training, such that the training data is enriched and more diverse.

\begin{figure*}[t!]
\centering
\centerline{\includegraphics[width=0.96\textwidth]{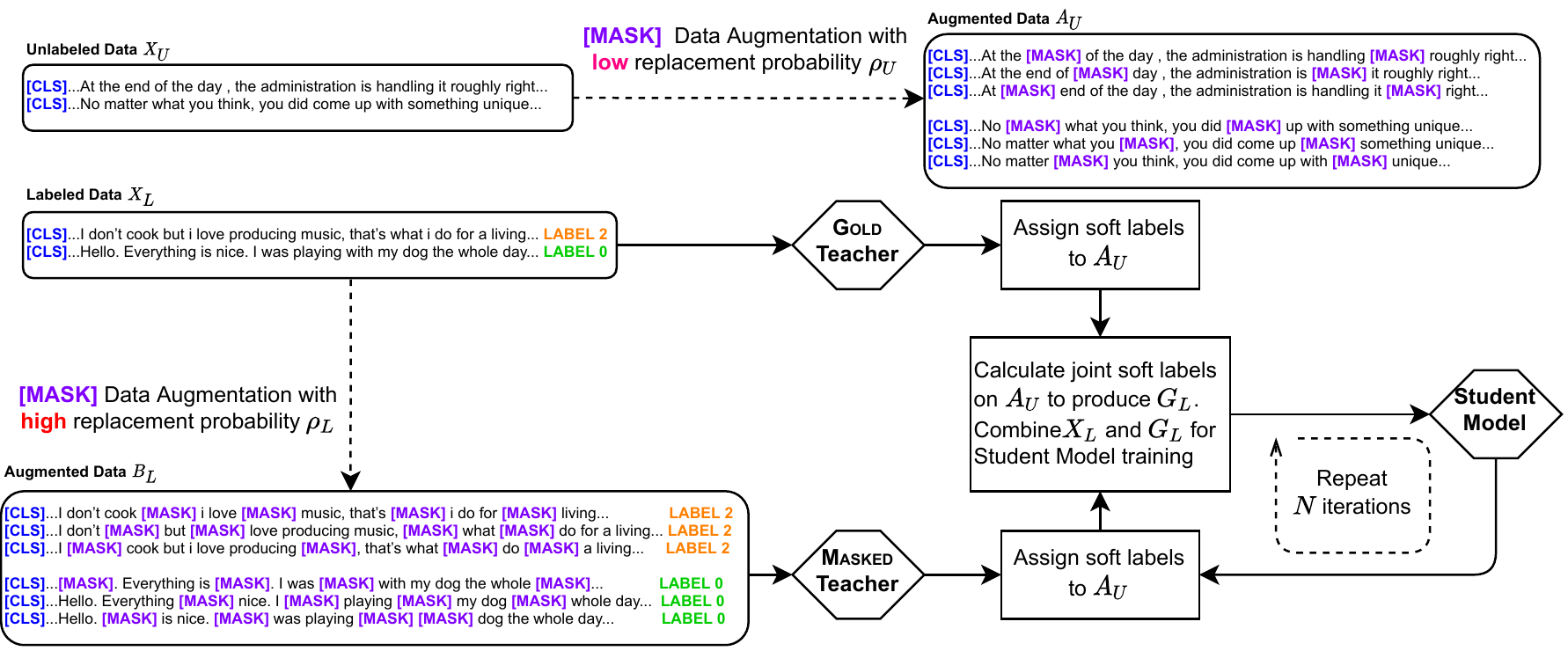}}
\caption{Overview of the proposed training process.}
\label{fig:train-overview}
\end{figure*}

In this work\footnote{\label{footnote:code-release}{The source code and trained models of this paper are available at https://github.com/nusnlp/S2T2.}}, we propose a novel semi-supervised teacher-student learning framework to improve the performance of pre-trained language models with unlabeled data.
We leverage unlabeled data from other sources to enrich the training set through self-training, which is a general case of domain adaptation where source data and target data are sampled from different data sources.
Self-training uses a trained classifier to assign label score vectors on unlabeled data instances.
However, such labeling process tends to generate labels under the assumption that a similar distribution is shared by labeled data and unlabeled data.
Since the distribution of unlabeled data is difficult to estimate, we introduce two teachers to improve the labeling of unlabeled data.
The student model is encouraged to integrate the knowledge from two teachers in a bootstrapping manner.

We leverage a data augmentation method~\citep{yavuz-etal-2020-simple} incorporating [MASK] tokens derived from a Masked Language Model (MLM) pre-training objective of pre-trained language models (PLM).
It is a natural fit to incorporate such augmented data with PLM like RoBERTa~\citep{liu-2019-roberta} and XLM-R~\citep{conneau-etal-2020-xlmr}, since the PLMs have adapted to the masking patterns during the pre-training process on large corpora.
The \textsc{Gold} teacher learns knowledge from only labeled data and it tends to generate labels following the distribution of labeled data.
The \textsc{Masked} teacher is trained with only perturbed labeled data which is augmented by randomly replacing tokens from the labeled data with the [MASK] tokens based on a predefined probability.

We construct the training data for the student model as the combination of two segments: labeled data and [MASK]-perturbed unlabeled data.
We explicitly impose a difference between masking probabilities applied to labeled data (for training the \textsc{Masked} Teacher) and unlabeled data (for training the student).
Two teachers can provide proper distribution estimation on these two data segments separately.
Therefore the student is optimized to distill the knowledge from the two teachers to improve self-training on the combined training set.

We evaluate our proposed approach on two multi-turn dialogue breakdown detection datasets and a large-scale follow-up question identification dataset.
Experimental results show that our semi-supervised teacher-student learning framework outperforms all previous published approaches and competitive supervised and semi-supervised baselines.
We also conduct further analysis to verify the effectiveness of the training strategies proposed in our framework.

\section{Task Overview}
Given a dialogue history $\mathcal{H}$ consisting of a sequence of alternating user and system utterances and the succeeding target system utterance $\mathcal{T}$, the task is to determine whether or not the target utterance causes a certain dialogue breakdown type.
Each instance ($\mathcal{H}, \mathcal{T}$) is associated with a soft label vector $\mathbf{y} \in \mathbb{R}^{|C|}$ which corresponds to the probability distribution over the set $C$ covering all possible breakdown types.

\section{Proposed Approach}
We give a detailed description of the proposed approach in this section.

\subsection{Pre-trained Language Model}
Assume that the dialogue history $\mathcal{H}=[\mathcal{H}_{1},...,\mathcal{H}_{h}]$ consists of $h$ tokens after tokenization and the target system utterance $\mathcal{T}=[\mathcal{T}_{1},...,\mathcal{T}_{t}]$ has $t$ tokens.
We first obtain a sequence of tokens by concatenating the two sequences with [CLS] and [SEP] tokens. The input to the pre-trained language model is:
\begin{equation}
    x=\mathrm{[CLS]}\mathcal{H}_{1}...\mathcal{H}_{h}\mathrm{[SEP]}\mathcal{T}_{1}...\mathcal{T}_{t}
\end{equation}

The combined sequence $x$ includes $n$ tokens.
$x$ is first converted to embedding $\mathbf{x}$ by the PLM embedding layer.
The output of the pre-trained language model is a sequence of hidden states from the last layer of the model:
\begin{equation}
    f(\mathbf{x};\theta_{m}) = \mathrm{PLM}(\mathbf{x})
\end{equation}
where $\theta_{m}$ denotes the parameters of PLM, $d$ is the hidden size of the pre-trained language model and the output shape is ${n \times d}$.

\subsection{Data Augmentation with [MASK] Tokens}
We leverage the pre-trained language models with Masked Language Model (MLM) training objective~\citep{devlin-etal-2019-bert,liu-2019-roberta,conneau-etal-2020-xlmr} for data augmentation with [MASK] tokens.

We perform data augmentation on the available data by randomly replacing the tokens in the instances with [MASK] tokens~\citep{yavuz-etal-2020-simple}.

For each labeled or unlabeled data instance, we generate a certain number of new data instances with [MASK] tokens based on a predefined replacement probability $\rho$. For instance, we replace 10\% of tokens with [MASK] tokens with $\rho=0.1$.

For the unlabeled data $X_U=\{(\mathbf{x}_{i}, \cdot)\}^{|X_U|}_{i=1}$, we eventually obtain the augmented data $\{(\mathbf{x}'_{j}, \cdot)\}^{k|X_U|}_{j=1}$, where $k$ is the number of augmented instances per original instance.
Similarly, for the labeled data $X_{L}=\{(\mathbf{x}_{i}, \mathbf{y}_{i})\}^{|X_L|}_{i=1}$, we obtain the augmented dataset $\{(\mathbf{x}'_{j}, \mathbf{y}'_{j})\}^{k|X_L|}_{j=1}$ where the label of the augmented instance remains the same as the original instance.
We use subscript $L$ to indicate that the dataset is labeled and use subscript $U$ for an unlabeled dataset.

\subsection{Teacher Models}
\label{sub:teacher-model-overview}
We introduce two teacher models, namely {Gold} Teacher (\textsc{gt}) and \textsc{Masked} Teacher (\textsc{mt}). Both teacher models share the same neural network architecture. Specifically, \textsc{Gold} Teacher is trained with labeled dataset (gold data) while \textsc{Masked} Teacher is trained with [MASK] augmented data.

A teacher model is formed by a pre-trained language model and a classification layer.
We denote parameters of the teacher model as $\theta^{(\textsc{t})}=\{\theta^{(\textsc{t})}_{m}, \theta^{(\textsc{t})}_{c}\}$ where $\theta^{(\textsc{t})}_{m}$ denotes parameters of the pre-trained language model and $\theta^{(\textsc{t})}_{c}$ the parameters of the classification layer.
We use the output at the first position ([CLS]) as the representation of the input $\mathbf{x}$:
\begin{equation}
\label{eqn:prtlm-cls-output}
    \mathbf{h} = f(\mathbf{x}; \theta^{(\textsc{t})}_{m})[0]
\end{equation}
The classification layer consists of two linear functions connected by $\mathrm{tanh}$ activation.
\begin{equation}
    \mathbf{g} = \mathbf{W}_{2}(\mathrm{tanh}(\mathbf{W}_{1}\mathbf{h} + \mathbf{b}_{1}))+\mathbf{b}_{2} 
\end{equation}
where $\mathbf{W}_{1} \in \mathbb{R}^{d \times d}$, $\mathbf{h},\mathbf{b}_{1} \in \mathbb{R}^{d}$, $\mathbf{W}_{2} \in \mathbb{R}^{|C| \times d}$ and $\mathbf{g}, \mathbf{b}_{2}  \in \mathbb{R}^{|C|}$.
The prediction is calculated by:
\begin{equation}
    \hat{\mathbf{y}} = f(\mathbf{x};{\theta^{(\textsc{t})}}) = \mathrm{softmax}(\mathbf{g})
\end{equation}
The loss function is a weighted sum of three objectives: cross-entropy loss $\mathcal{L}_{\mathrm{CE}}$, supervised contrastive learning loss $\mathcal{L}_{\mathrm{SCL}}$~\citep{gunel2021supervised}, and mean squared error loss $\mathcal{L}_{\mathrm{MSE}}$.
\begin{equation}
\label{eqn:loss}
    \mathcal{L} = \beta_{1}\mathcal{L}_{\mathrm{CE}} + \beta_{2}\mathcal{L}_{\mathrm{SCL}} + \beta_{3}\mathcal{L}_{\mathrm{MSE}}
\end{equation}

The supervised contrastive learning loss is defined as:
\begin{equation}
\begin{aligned}
    \label{eqn:loss_SCL}
    \mathcal{L}_{\mathrm{SCL}} =&\sum_{i=1}^{N_{b}} -\frac{1}{N_{b,y_i}-1} \sum_{j=1}^{N_{b}} \mathds{1}_{i \neq j} \mathds{1}_{y_i = y_j}  \\
    \log& \frac{\exp{(\Phi(\mathbf{x}_i}) \cdot \Phi(\mathbf{x}_j) / \tau)}{\sum_{k=1}^{N_{b}} \mathds{1}_{i \neq k} \exp{(\Phi(\mathbf{x}_i) \cdot \Phi(\mathbf{x}_k)/ \tau)}}
\end{aligned}
\end{equation}
where $y_i=\mathrm{argmax}(\mathbf{y}_i)$, $N_{b}$ is the batch size, and $N_{b,y_i}$ is the number of instances with the same label as the $i$-th instance within the batch. $\mathds{1}$ denotes the indicator function. $\tau$ is a temperature parameter. $\Phi(\cdot)$ corresponds to the encoder function described in Eqn.~\ref{eqn:prtlm-cls-output}.

We fine-tune all parameters in $\theta^{(\textsc{t})}$. We use $\theta^{(\textsc{gt})}$ and $\theta^{(\textsc{mt})}$ for {Gold} Teacher and \textsc{Masked} Teacher respectively.

\subsubsection{\textsc{Gold} Teacher}
\label{subsub:gold-teacher}
The \textsc{Gold} Teacher is fine-tuned on labeled dataset $X_{L}=\{(\mathbf{x}_{i}, \mathbf{y}_{i})\}^{|X_L|}_{i=1}$. It learns knowledge purely from quality annotated data.

Given the unlabeled dataset $X_U=\{(\mathbf{x}_{i}, \cdot)\}^{|X_U|}_{i=1}$, we augment $X_U$ to $A_U=\{(\mathbf{x}'_{i}, \cdot)\}^{k|X_U|}_{i=1}$ with $k$ augmented instances per original unlabeled instance and the [MASK] token replacement probability $\rho_{U}$.

We use the fine-tuned \textsc{Gold} Teacher to assign soft labels to the augmented unlabeled dataset $A_U$.

\subsubsection{\textsc{Masked} Teacher}
\label{subsub:mask-teacher}
Given the labeled dataset $X_{L}=\{(\mathbf{x}_{i}, \mathbf{y}_{i})\}^{|X_L|}_{i=1}$ which is the training data for \textsc{Gold} Teacher, we prepare the training data for \textsc{Masked} Teacher by generating an augmented dataset on $X_L$.

The [MASK] augmentation on $X_L$ is determined by $k$ and the [MASK] token replacement probability $\rho_{L}$ which results in the augmented dataset $B_L=\{(\mathbf{x}'_{i}, \mathbf{y}'_{i})\}^{k|X_L|}_{i=1}$. The \textsc{Masked} Teacher is fine-tuned on $B_L$.

The \textsc{Masked} Teacher learns from training data with [MASK] tokens which adapts to the situation of predicting labels on instances with [MASK] tokens.

We use the fine-tuned \textsc{Masked} Teacher to assign soft labels to unlabeled instances from $A_U$ which is the same augmented unlabeled dataset described for Gold Teacher.

We set [MASK] token replacement probability $\rho_{L}$ to be larger than $\rho_{U}$ ($\rho_{L}>\rho_{U}$) such that the \textsc{Masked} Teacher is robust to produce more confident label scores on $A_U$.

\subsection{Student Model}
\label{subsec:student-model}
The student model follows the same architecture as the teacher model, consisting of a pre-trained language model and a classification layer.

The parameters of the student model are denoted as $\theta^{(\textsc{s})}=\{\theta^{(\textsc{s})}_{m}, \theta^{(\textsc{s})}_{c}\}$ which correspond to the pre-trained language model and the classification layer. The pre-trained language model inherits the weights from the \textsc{Gold} Teacher fine-tuned on $X_L$, that is, $\theta^{(\textsc{s})}_{m} := \theta^{(\textsc{gt})}_{m}$.
We do not perform fine-tuning of $\theta^{(\textsc{s})}_{m}$ during the training of the student model.

The training objective of the student model is the same as the teacher model, which is defined in Eqn.~\ref{eqn:loss}.

\section{Training Process}
\label{sec:train-process}
As mentioned in the last section, the \textsc{Gold} Teacher is fine-tuned on the labeled dataset $X_L$ and the \textsc{Masked} Teacher is fine-tuned on $B_L$ where $B_L$ is augmented based on $X_L$. We also have the unlabeled dataset $A_U$ which is augmented based on the unlabeled dataset $X_U$.
We present the overall training process in Figure~\ref{fig:train-overview}.
\subsection{Joint Scoring on Unlabeled Data}
\label{sub:score-udata}
For each $\mathbf{x}'\in A_U$, we assign label score vectors by both the \textsc{Gold} Teacher and the \textsc{Masked} Teacher respectively.
\begin{align}
    \hat{\mathbf{y}}_{\textsc{gt}}&=f(\mathbf{x}';{\theta^{(\textsc{gt})}}) \\
    \hat{\mathbf{y}}_{\textsc{mt}}&=f(\mathbf{x}';{\theta^{(\textsc{mt})}})
\end{align}

We define a joint scoring function to compute the label score vector with weights determined by hyperparameter $\gamma$.
\begin{equation}
\label{eqn:joint-scoring}
    \hat{\mathbf{y}} = \gamma\hat{\mathbf{y}}_{\textsc{gt}} + (1-\gamma)\hat{\mathbf{y}}_{\textsc{mt}}
\end{equation}
We then obtain the labeled set $G_L=\{(\mathbf{x}'_{i}, \mathbf{y}'_{i})\}^{|A_U|}_{i=1}$ where $\mathbf{x}'\in A_U$ and $\mathbf{y}'$ is calculated by Eqn.~\ref{eqn:joint-scoring}.

\subsection{Bootstrapping Strategy}
We consider a bootstrapping strategy to refine the labeling of unlabeled data to improve classification.
Continuing with $G_L$ obtained by the process mentioned in the last subsection, we describe the bootstrapping strategy for student model fine-tuning as follows.

We use the combined dataset $X_{L}^{(\textsc{s})}=X_L\cup G_L$ as the initial training set for the student model. After a complete training iteration consisting of $k_{e}$ epochs, the fine-tuned student model predicts label score vectors for each unlabeled data $\mathbf{x}'\in A_U$.
\begin{equation}
    \hat{\mathbf{y}}_{\textsc{s}}=f(\mathbf{x}';{\theta^{(\textsc{s})}})
\end{equation}
The refined label score vector after each training iteration is calculated by:
\begin{equation}
\label{eqn:bootstrap-update-score}
    \begin{aligned}
    \lambda &= i/N \quad (i=1,2,...,N-1) \\
    \hat{\mathbf{y}} &= \alpha[(1+\lambda)\hat{\mathbf{y}}_{\textsc{s}} + (1-\lambda)\hat{\mathbf{y}}_{\textsc{mt}}]
    \end{aligned}
\end{equation}
where $i$ is the iteration index and $N$ the total number of iterations.
Therefore, $\lambda$ ranges between 0 and 1 ($0<\lambda<1$).
We set $\alpha=0.5$ in our experiments, such that we ramp up the weight of $\hat{\mathbf{y}}_{\textsc{s}}$ from 0.5 to 1.0 while progressively decreasing the contribution from the \textsc{Masked} Teacher to produce better predictions for the unlabeled data.

As a result, the label score vectors in $G_L$ and the combined training set $X_{L}^{(\textsc{s})}$ are updated after each training iteration.

For training of the student model in the succeeding iteration, we retain the parameters from the best epoch in the last iteration based on development set performance.
We use the trained student model to make predictions on test sets.

\begin{table}[tb!]
\small
\centering
\begin{tabular}{l|c}
\hline
\textbf{English} & Train / Dev / Unlabeled / Test \\
\hline
\#instances	& 2,110 / 1,950 / 6,150 / 1,940 \\\hline
\end{tabular}
\begin{tabular}{l|c}
\hline
\textbf{Japanese} & Train / Dev / Unlabeled / Test \\
\hline
\#instances	& 10,418 / 2,818 / 11,506 / 2,672 \\\hline
\end{tabular}
\caption{Statistics of DBDC5 English and Japanese datasets.}
\label{tab:dbdc-data}
\end{table}

\begin{table}[tb!]
\small
\centering
\begin{tabular}{l|cccc}
\hline
\textbf{LIF} & Train/Dev/Test-I/Test-II/Test-III\\
\hline
\#instances	& 126,632/5,861/5,992/5,247/2,685 \\\hline
\multicolumn{2}{c}{\#unlabeled~~~~~101,448} \\\hline
\end{tabular}
\caption{Statistics of LIF dataset.}
\label{tab:lif-data}
\end{table}

\section{Experiments}
\subsection{Datasets}
We evaluate our proposed approach on two multi-turn dialogue datasets DBDC5 English Track~\citep{higashinaka-2020-dbdc5-overview} and DBDC5 Japanese Track~\citep{higashinaka-2020-dbdc5-overview}\citep{higashinaka-2020-dbdc5-overview}, and one much larger Learning to Identify Follow-Up Questions dataset LIF~\citep{kundu-etal-2020-learning}.

\noindent\textbf{DBDC5 English Track} This is a multi-turn dialogue dataset which requires identification of the predefined dialogue breakdown type of the last system utterance given the dialogue history.
Based on the annotation quality~\citep{higashinaka-2020-dbdc5-overview}, we use the re-annotated DBDC4 data as the labeled dataset.
For the unlabeled data, we use the English data released in ~\citet{dbdc3-overview-2017}.

\noindent\textbf{DBDC5 Japanese Track} This is a Japanese dataset with the same format as DBDC5 English Track. For the Japanese track, we use datasets released in previous DBDC tasks as training set, including DBDC1, DBDC2, DBDC3, and DBDC4 development sets, as well as DBDC5 development set.
We use DBDC4 evaluation set for validation. These data were annotated by 15--30 annotators per instance.
We use Chat dialogue corpus as the source of unlabeled data, which were annotated by only 2--3 annotators.~\citep{higashinaka-2019-dbdc4-overview}

\noindent\textbf{LIF} LIF~\citep{kundu-etal-2020-learning} is a conversational question answering dataset for the task of follow-up question identification, which requires the model to identify whether or not the last question follows up on the context passage and previous conversation history.
Since LIF is derived from QuAC~\citep{choi-2018-quac}, we select the training set of CoQA~\citep{reddy-2019-coqa} which is a similar conversational QA dataset, as the source of unlabeled data.\footnote{\label{footnote:coqa}{As we use CoQA samples without modification, the samples do not include the cases where the candidate question is from other conversations~\citep{kundu-etal-2020-learning}, we suggest these samples still contribute to the generalization.}}

We present the statistics of both DBDC5 datasets in Table~\ref{tab:dbdc-data} and the statistics of the LIF dataset in Table~\ref{tab:lif-data}. The numbers reported do not include augmented data.

\subsection{Evaluation Metrics}
DBDC5 English Track and DBDC5 Japanese Track require classification-based metrics including accuracy and F1 scores, and distribution-based metrics including Jensen-Shannon divergence (JSD) and Mean Squared Error (MSE).\footnote{\label{footnote:dbdc-metrics-ref}{Refer to \citet{higashinaka-2020-dbdc5-overview} for details.}}

LIF dataset requires classification-based metrics including precision, recall, and F1 of class \textit{Valid}, and macro F1.\footnote{\label{footnote:lif-metrics-ref}{Refer to \citet{kundu-etal-2020-learning} for details.}}

\begin{table*}[tb!]
\small
\centering
\begin{tabular}{l|cccc|cccc}
\hline
&  \multicolumn{4}{c|}{\textbf{English}} & \multicolumn{4}{c}{\textbf{Japanese}}\\
Model & Accuracy & F1(B) & JSD$\downarrow$ & MSE$\downarrow$
& Accuracy & F1(B) & JSD$\downarrow$ & MSE$\downarrow$ \\
\hline \hline
BERT+SSMBA
& 0.739	& 0.782	& 0.070  & 0.036
& --  & --  & --  & -- \\
XLMR+CM
& --  & --  & --  & --
& 0.745	& 0.694	& 0.077  & 0.040\\
PLM$_{\flat}$ Baseline$^{\#}$
& 0.721 & 0.765 &0.080 & 0.043
& 0.706 & 0.659 &0.084 & 0.044 \\
PLM Baseline
& 0.750 & 0.797 & 0.066 & 0.033
& 0.732 & 0.650 & 0.074 & 0.039 \\
PLM+CoAtt
& 0.752 & 0.794 & 0.067 & 0.033
& 0.740 & 0.708 & 0.069 & 0.035 \\
RoBERTa+SSMBA & 0.752  & 0.797 & 0.067 & 0.035 & -- & -- & -- & -- \\
UDA & 0.754	& 0.799 & 0.071 & 0.037 & 0.733 & 0.692 & 0.075 & 0.039 \\
MixText & 0.757 & 0.805 & 0.059 & 0.030  & 0.743 & 0.715 & 0.073 & 0.036\\\hline
Ours ($X_U$, no $A_U$) & 0.759	& 0.803 & 0.065 & 0.033 & 0.747	& 0.721 & 0.068 & 0.036
\\
Ours   
& \bf{0.779}	& \bf{0.824}	& \bf{0.058}  & \bf{0.028}
& \bf{0.767}	& \bf{0.754}	& \bf{0.062}  & \bf{0.031}
\\
\hline
\end{tabular}
\caption{Experimental results on the DBDC5 English and Japanese track. $\downarrow$ the lower the better. $\#$ subscript $\flat$ denotes the base version of PLM.}
\label{tab:dbdc-results}
\end{table*}

\begin{table*}[thb!]
\small
\centering
\begin{tabular}{l|c|c|c}
\hline
& \textbf{Test-I}  & \textbf{Test-II}  & \textbf{Test-III} \\
\textbf{Models}  & V-P/-R/-F1/Macro F1 & V-P/-R/-F1/Macro F1 & V-P/-R/-F1/Macro F1 \\ \hline \hline
Three-way AP    & 74.4/75.7/75.0/81.4 & 89.0/75.7/81.8/86.2 & 81.9/75.7/78.7/65.0 \\
PLM Baseline & 75.6/85.4/80.2/84.9   & 88.2/85.4/86.8/89.6 & 84.2/85.4/84.8/72.0 \\
PLM+CoAtt & 76.6/80.6/78.5/83.9   & 87.7/80.6/84.0/87.6   & 85.8/80.6/83.1/71.8\\
UDA & \textbf{79.2}/83.9/81.5/86.1  & \textbf{91.4}/83.9/87.5/90.3 & 85.6/83.9/84.7/73.2 \\
MixText & 74.8/86.6/80.3/84.8  & 87.8/86.6/87.2/89.9  & 83.4/86.6/85.0/71.5 \\\hline
Ours ($X_U$, no $A_U$) & 78.0/83.9/80.9/85.6		& 89.5/83.9/86.6/89.6		& \textbf{85.9}/83.9/84.9/73.5
\\
Ours (4.74$\%^{\star}$)  & 73.9/83.3/78.3/83.5  & 87.5/83.3/85.4/88.6  & 82.6/83.3/83.0/69.0 \\
Ours (25$\%^{\star}$)    & 74.6/86.9/80.3/84.8		& 88.7/86.9/87.8/90.4	& 82.4/86.9/84.6/70.1 \\
Ours (50$\%^{\star}$)    & 77.6/\textbf{87.3}/82.1/86.4  & 89.3/\textbf{87.3}/88.3/90.8  & 85.5/\textbf{87.3}/\textbf{86.4}/\textbf{74.8} \\
Ours (100$\%^{\star}$) & 78.1/86.6/\textbf{82.2}/\textbf{86.5} & 90.6/86.6/\textbf{88.6}/\textbf{91.0} & 85.0/86.6/85.8/73.8 \\
\hline
\end{tabular}
\caption{Experimental results on the LIF dataset. V-P, V-R, and V-F1 correspond to precision, recall, and F1 score on class \textit{Valid}. $\star$ denotes the percentage of the LIF training dataset used.}
\label{tab:lif-result}
\end{table*}

\subsection{Experimental Setup}
We experiment with RoBERTa~\citep{liu-2019-roberta} as the pre-trained language model in DBDC5 English Track and LIF. We use multilingual pre-trained language model XLM-R~\citep{conneau-etal-2020-xlmr} for DBDC5 Japanese Track. We use the large version with hidden size $d = 1024$.
The maximum input length is set to 256.

For experiments on LIF, we concatenate the context passage and the conversation history into $\mathcal{H}$, and $\mathcal{T}$ corresponds to the candidate question.

The weights $\beta_{1}$, $\beta_{2}$, $\beta_{3}$ are set to 1e-2, 1e-3, and 1.0 in the loss function for both DBDC5 English Track and Japanese Track.
Since LIF does not require distribution-based metrics, the weights $\beta_{1}$, $\beta_{2}$, $\beta_{3}$ are set to $1.0$, $0.1$, and $0$ in experiments on LIF.
We set temperature $\tau$ to $1.0$ in $\mathcal{L}_{\mathrm{SCL}}$ and $\gamma$ to $0.5$ in Eqn.~\ref{eqn:joint-scoring}.
We optimize the loss using AdamW~\citep{loshchilov2018decoupled} with $0.01$ weight decay.

For data augmentation, we generate 6 instances for each labeled or unlabeled instance.
We set [MASK] token replacement probability $\rho_{U}=0.15$ aligning to \citep{devlin-etal-2019-bert,liu-2019-roberta} and $\rho_{L}=0.25$.

To train two teacher models, we use a batch size of 16, 8, and 12 for experiments on DBDC5 English Track, DBDC5 Japanese Track, and LIF respectively. The learning rate during training is set to 1e-5, 1e-5, and 2e-6 respectively.
To train the student model, we use a batch size of 128 and learning rate 2e-6 for experiments on all three datasets.
We set the maximum number of iterations $N$ to 5 and the number of epochs $k_e$ to 5 per iteration.
Models are trained on a single Tesla V100 GPU.

\subsection{Compared Models}
\textbf{BERT+SSMBA}~\citep{ng-2020-improving}
The model consists of pre-trained language model BERT-base and a classification layer. The BERT parameters are further pretrained on large-scale Reddit dataset.
The labeled training data is augmented based on
SSMBA~\citep{ng-etal-2020-ssmba} and original labels are assigned to augmented instances.
This is the best-performing model published to date on the DBDC5 English track.
We implement a baseline \textbf{RoBERTa+SSMBA} using RoBERTa classification model with SSMBA augmentation.
For fairer comparison with our proposed approach on DBDC5 English dataset, we adopt RoBERTa-large model and generate data with SSMBA which follows BERT+SSMBA.

\noindent\textbf{XLMR+CM}
The model uses cross-lingual language model XLM-R with context matching (CM) modules.
This is the best-performing model published to date on the DBDC5 Japanese track~\citep{higashinaka-2020-dbdc5-overview}.

\noindent\textbf{Three-way AP}~\citep{kundu-etal-2020-learning}
The model applies an attentive pooling network to capture interactions among the context passage, conversation history, and the candidate follow-up question. This is the best performing-model published to date on the LIF dataset.

\noindent\textbf{PLM Baseline}
We build a simple but effective baseline model consisting of a pre-trained language model and a classification layer.
We select RoBERTa for experiments on English tasks (DBDC5 English Track and LIF) and XLM-R for experiments on DBDC5 Japanese track.
We use the output from [CLS] as the representation for classification.
We adopt the large version of the pre-trained language model (RoBERTa-large or XLM-R-large) unless stated otherwise.

\noindent\textbf{PLM+CoAtt}
We build another baseline by applying a co-attention network~\citep{lin-etal-2020-co} on the output from a pre-trained language model.
The selection of pre-trained language models follows \textbf{PLM Baseline}.
Since the co-attention network applies to two sequences of representations corresponding to conversation history and the last utterance, we prepend the context passage to the conversation history and the candidate question is treated as the last utterance for experiments on LIF.

We also experiment with recently proposed semi-supervised methods \textbf{UDA}~\citep{xie-2020-uda} and \textbf{MixText}~\citep{chen-etal-2020-mixtext}.
We adopt RoBERTa-large for the English datasets and XLM-R-large for the Japanese dataset.
Labeled and unlabeled data (before augmentation) used are the same as our proposed method.
For English datasets DBDC5 English and LIF, we use back-translation with German and Russian as intermediate languages for augmentation on unlabeled data following~\citet{chen-etal-2020-mixtext}.
For DBDC5 Japanese dataset, we apply [MASK] augmentation used in our proposed method due to the non-availability of Japanese round-trip back-translation model.
UDA and MixText are trained on both labeled and unlabeled data, while the other compared models are trained on labeled data in a supervised manner.

\section{Results}
\subsection{Main Results}
We present the experimental results of DBDC5 (both English Track and Japanese Track) and LIF in Table~\ref{tab:dbdc-results} and Table~\ref{tab:lif-result}, respectively.
Results of BERT+SSMBA and XLMR+CM are retrieved from \citet{higashinaka-2020-dbdc5-overview} and results of Three-way AP are retrieved from \citet{kundu-etal-2020-learning}.
For results of both DBDC5 datasets, we report Accuracy, F1(B), JS Divergence (JSD), and Mean Squared Error (MSE).
For metrics MSE and JSD, the reported percentage of improvement is calculated as $100 - (\textit{ours}/\textit{other\_model})\times 100$.

In the DBDC5 English Track, our proposed approach outperforms the prior best-performing model (BERT+SSMBA) by 4.0\%, 4.2\%, 17.1\%, and 22.2\% on metrics Accuracy, F1(B), JSD, and MSE.
It also performs better than all supervised and semi-supervised baselines by at least 2.2\%, 1.9\%, 1.7\%, and 6.7\% on the reported four metrics.
The results of RoBERTa+SSMBA show that based on the large pre-trained language model setting, adding SSMBA augmented data does not contribute improvement on this task.

In the DBDC5 Japanses Track, our proposed approach outperforms the prior best-performing model (XLMR+CM) by 2.2\%, 6.0\%, 19.5\%, and 22.5\% on metrics Accuracy, F1(B), JSD, and MSE.
The improvements are at least 2.4\%, 3.9\%, 10.1\%, and 11.4\% when compared to all other baseline models.
We notice that models with large version of PLM perform generally better on these datasets.

In the much larger LIF dataset, we sample different sizes of labeled training data from the full training dataset to verify the robustness of our approach.
The smallest sampled training set consists of only 6,000 (4.74\% of 126,632) labeled training instances, similar to the sizes of Test-I and Test-II.
In this case, we sample 18,000 instances from CoQA as unlabeled data and increase the sample size accordingly for the larger training sets.
With only 6,000 labeled training instances, our method achieves competitive performance which outperforms the previous best-performing model (Three-way AP) on all three LIF test sets except V-P of Test-I and Test-II.
We also experiment with 25\%, 50\%, and 100\% of full training data and observe further performance improvement.
With 100\% labeled training data, our approach outperforms Three-way AP on all metrics by a wide margin and also performs better than other supervised and semi-supervised baselines on all metrics except for V-P.

\begin{table}[tb!]
\small
\centering
\begin{tabular}{lccc}
\hline
\textbf{Model} & D-EN & D-JP & LIF \\
\hline
Ours (full)	& \textbf{77.9} & \textbf{76.7} & \textbf{86.5} \\\hline
-- MT  & 76.9 & 75.9 & 86.1 \\
~~~-- self-training    & 75.0 & 73.2 & 84.9 \\
-- GT  & 74.0 & 73.4 & 85.9 \\
~~~-- self-training   & 73.8 & 73.1 & 84.8 \\\hline
\end{tabular}
\caption{Performance on the test set after removing (--) different components. We report Accuracy on DBDC5 English (D-EN) and DBDC5 Japanese (D-JP) and Macro F1 on LIF Test-I. GT: \textsc{Gold} Teacher. MT: \textsc{Masked} Teacher.}
\label{tab:ablation-GTMT}
\end{table}

We conduct experiments comparing the use of $X_U$ and $A_U$ when training the student model.
We replace $A_U$ as $X_U$ in our original approach and denote this variation as Ours ($X_U$, no $A_U$).
The results on the test sets indicate that utilizing $A_U$ (augmentations on unlabeled data) is more effective.

We perform statistical significance tests with regards to Accuracy (DBDC5 English and Japanese datasets) and Macro F1 (LIF) on test sets.
Our proposed method is significantly better ($p<0.05$) than all baseline methods.

\subsection{Analysis}
\label{sub:analysis}
Based on our implementation of teacher and student models where teachers and student use the same architecture, our proposed approach is in line with the idea of self-distillation~\citep{mobahi-2020-self,furlanello-2018-born-again}.
It has been observed that self-distillation helps to improve test performance~\citep{liu-etal-2021-noisy,furlanello-2018-born-again,zhang-2019-own-teacher}.
\citet{zhu-2020-towards} show that self-distillation performs implicit ensemble with knowledge distillation.
In traditional self-distillation, the student is distilled from a single trained teacher.
In our approach, we distill the knowledge from two different trained teachers with additional unlabeled data.
We evaluate the effectiveness of two teachers by removing components in the proposed approach and show the results in Table~\ref{tab:ablation-GTMT}.
Performance drops when we remove either teacher but keep the self-training on combined training set.
We observe that performance drops further if we continue to remove the self-training process.
This shows that both teachers contribute to the performance improvement on these tasks.

We conduct analysis of different settings on the development set, in order to select hyperparameters as well as to better understand the effectiveness of our proposed approach.
For development set performance, we report accuracy score in the DBDC5 English Track and DBDC5 Japanese Track, and Macro F1 score in LIF6000 in which the model is provided with 6,000 labeled training instances sampled from LIF.

We select the number of augmented samples $k$ from \{4,6,8\} and observe that the sample size 6 performs consistently better than the other two on all three datasets.

We investigate the impact of [MASK] token replacement probability of labeled data ($\rho_{L}$) on the model performance. Given a constant $\rho_{U}=0.15$, we vary the value of $\rho_{L}$ from $0.15$ to $0.30$ with step size of $0.05$.
The best development set performance is achieved at $\rho_{L}=0.25$.

\begin{figure}[tb!]
    \centering 
    \subfigure{\includegraphics[width=0.32\textwidth]{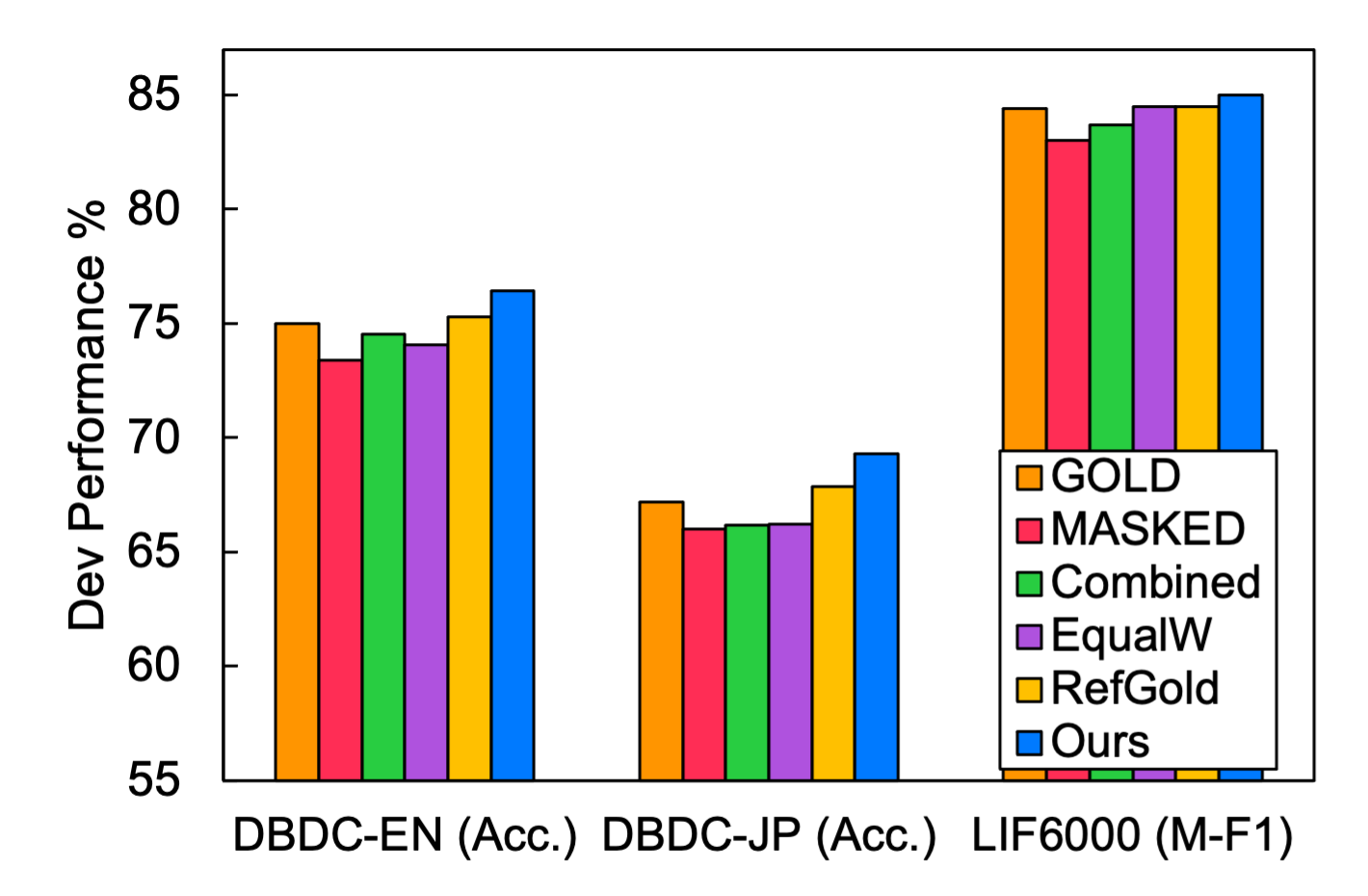}\label{fig:train-strategies}}
    \caption{Performance comparison on different training strategies on the development set.}
    \label{fig:ablation}
\end{figure}

We also explore different training strategies and compare them with our proposed approach.
\textbf{\textsc{Gold}} denotes that we only use \textsc{Gold} Teacher which is trained on labeled data $X_{L}$ only to make predictions.
\textbf{\textsc{Masked}} denotes that we only use \textsc{Masked} Teacher which is trained on [MASK] augmentation of labeled data $B_{L}$ to make predictions.
\textbf{Combined} means the model is trained from scratch on the combination of labeled data and [MASK] augmentation of labeled data ($X_{L}\cup B_{L}$) without bootstrapping.
\textbf{EqualW} indicates the training method in which we use trained \textsc{Gold} Teacher and trained \textsc{Masked} Teacher to make predictions on unlabeled data and score equally for the final label scores.
That is, $\hat{\mathbf{y}} = 0.5\hat{\mathbf{y}}_{\textsc{gt}} + 0.5\hat{\mathbf{y}}_{\textsc{mt}}$ in Eqn.~\ref{eqn:joint-scoring} for generating labels for unlabeled data and obtaining $G_L$.
We then train the student model on $X_{L}\cup G_L$ without bootstrapping.
\textbf{RefGold} denotes a variant of our bootstrapping approach where the score refinement in Eqn.~\ref{eqn:bootstrap-update-score} is altered to $\hat{\mathbf{y}} = \alpha[(1+\lambda)\hat{\mathbf{y}}_{\textsc{s}} + (1-\lambda)\hat{\mathbf{y}}_{\textsc{gt}}]$ in which we refer to \textsc{Gold} Teacher. We present performance comparison in Figure~\ref{fig:ablation}.

Our proposed approach outperforms all other mentioned training strategies on the development set.
\textbf{\textsc{Gold}} achieves the best performance among non-bootstrapping settings, indicating that preserving knowledge from \textbf{\textsc{Gold}} Teacher is important, which validates the initialization of our proposed student model.
\textbf{RefGold} produces slightly lower scores than our proposed approach, probably because the number of masked training instances is more than instances without [MASK] tokens during bootstrapping training, so using predictions (Eqn.~\ref{eqn:bootstrap-update-score}) from \textsc{Masked} Teacher is better.
But it still performs better than other non-bootstrapping settings. This finding suggests that the proposed bootstrapping is essential for further performance improvement.
Our proposed method (Ours) is also significantly better ($p<0.05$) than \textbf{\textsc{Gold}} and \textbf{RefGold}.

\section{Related Work}
\noindent\textbf{Data Augmentation}
Recent unsupervised data augmentation methods have shown the effectiveness on classification tasks with short text instances.
\citet{wei-zou-2019-eda} introduce random word-level operations including replacement, insertion, deletion, and swapping.
\citet{xie-2020-uda} add noise to the unlabeled data and generate new training data by back-translation.
These augmentation methods tend to generate unnatural text samples as the text sequence becomes longer such as multi-turn dialogues and conversations.
A recent data augmentation method based on self-supervised learning is proposed to tackle the out-of-domain issue~\citep{ng-2020-improving}.
Another line of recent work proposes to augment data by randomly replacing word tokens with [MASK] tokens while working with pre-trained language models pre-trained with Masked Language Model objective~\citep{yavuz-etal-2020-simple}.
In our work, we leverage this idea and further investigate how different probabilities of [MASK] token replacement affect the model performance.

\noindent\textbf{Domain Adaptation}
Domain adaptation is a general method which transfers the knowledge from a source domain to a target domain.
Domain adaptation is usually applied to text classification tasks when labeled source domain data is more abundant than target domain data.
It enables a classifier trained on a source domain to be generalized to another target domain~\citep{jiang-2007-domain,chen-2011-domain,chen-2012-domain}.
Recent works incorporate output features from pre-trained language models to improve domain adaptation~\citep{nishida-etal-2020-domain,ye-etal-2020-domain}.
In our work, we sample unlabeled data from sources other than the available labeled training set to enrich the training data.
We leverage the idea of domain adaptation with source data and target data sampled from different data sources.

\noindent\textbf{Semi-supervised Learning}
In our work, we utilize unlabeled data from other sources for model training via semi-supervised learning.
Semi-supervised learning involves both labeled and unlabeled data during training.
The general idea is to train a model with labeled data in a supervised learning manner and then enrich the labeled set with the most confident predictions on unlabeled data~\citep{kehler-etal-2004-competitive-self,mcclosky-etal-2006-self-training,oliver-2018-semi,li-2019-self-train}.
Regularization techniques are applied to obtain better decision boundaries of unlabeled data with unknown distribution, including adversarial training~\citep{miyato-2017-adversarial}, adding dropout, adding noise, and bootstrapping~\citep{laine-2017-semi}.
We consider the bootstrapping strategy to refine the labeling of unlabeled data.
Prior work shows that bootstrapping improves the labeling of unlabeled data~\citep{reed-2015-bootstrapping,laine-2017-semi,he-etal-2018-adaptive}.

\section{Conclusion}
In this work, we propose a novel semi-supervised teacher-student learning framework with two teachers.
We leverage both labeled and unlabeled data during training in a bootstrapping manner.
We show that bootstrapping with the proposed re-labeling method is essential to improve performance.
Evaluation results on two multi-turn dialogue breakdown detection datasets and a large-scale follow-up question identification dataset show that our proposed method achieves substantial improvements over prior published methods and competitive baselines.

\section*{Acknowledgements}

This research is supported by the National Research Foundation, Singapore under its AI Singapore Programme (AISG Award No: AISG-RP-2018-007). The computational work for this article was partially performed on resources of the National Supercomputing Centre, Singapore (https://www.nscc.sg).

\bibliography{aaai22}

\end{document}